\documentclass[10pt,twocolumn,letterpaper]{article}

\usepackage{cvpr}
\usepackage{times}
\usepackage{epsfig}
\usepackage{graphicx}
\usepackage{amsmath}
\usepackage{amssymb}
\usepackage{array}
\usepackage{booktabs}
\usepackage{mathrsfs}
\usepackage{algorithm}
\usepackage{algorithmic}
\usepackage{multirow}
\usepackage{mathtools}

\newcommand{\INPUT}{\item[\myinput]}
\newcommand{\myinput}{\textbf{Initialization:}}

\newcommand{\MYWHILE}{\item[\mywhile]}
\newcommand{\mywhile}{\textbf{repeat}}

\newcommand{\MYENDWHILE}{\item[\myendwhile]}
\newcommand{\myendwhile}{\textbf{until}}

\usepackage[pagebackref=true,breaklinks=true,letterpaper=true,colorlinks,bookmarks=false]{hyperref}

\cvprfinalcopy 

\def\cvprPaperID{447} 

\ifcvprfinal\fi
\begin{document}

\title{Incorporating Structural Alternatives and Sharing into Hierarchy for Multiclass Object Recognition and Detection}

\author{Xiaolong Wang$^1$, ~Liang Lin$^{1}$\thanks{Corresponding author is Liang Lin (linliang@ieee.org). This work was supported by National Natural Science Foundation of China (no. 61173082), the Special Project on the Integration of Industry, Education and Research of Guangdong Province (no. 2012B091100148), and the Guangdong Natural Science Foundation (no.S2011010001378). }, ~Lichao Huang$^1$, ~Shuicheng Yan$^2$\\
{\small $^1$Sun Yat-Sen University, Guangzhou, China}\\
{\small $^2$Department of ECE, National University of Singapore, Singapore}
}

\maketitle


\vspace{-4mm}
\begin{abstract}

\vspace{-4mm}

This paper proposes a reconfigurable model to recognize and detect multiclass (or multiview) objects with large variation in appearance. Compared with well acknowledged hierarchical models, we study two advanced capabilities in hierarchy for object modeling: (i)``switch'' variables(i.e. or-nodes) for specifying alternative compositions, and (ii) making local classifiers (i.e. leaf-nodes) shared among different classes. These capabilities enable us to account well for structural variabilities while preserving the model compact. Our model, in the form of an And-Or Graph, comprises four layers: a batch of leaf-nodes with collaborative edges in bottom for localizing object parts; the or-nodes over bottom to activate their children leaf-nodes; the and-nodes to classify objects as a whole; one root-node on the top for switching multiclass classification, which is also an or-node.  For model training, we present an EM-type algorithm, namely dynamical structural optimization (DSO), to iteratively determine the structural configuration, (e.g., leaf-node generation associated with their parent or-nodes and shared across other classes), along with optimizing multi-layer parameters. The proposed method is valid on challenging databases, e.g., PASCAL VOC 2007 and UIUC-People, and it achieves state-of-the-arts performance.

\end{abstract}

\vspace{-4mm}

\setlength{\abovedisplayskip}{3pt}
\setlength{\belowdisplayskip}{3pt}

\section{Introduction}

\begin{figure}[!htb]
\centering
\epsfig{figure=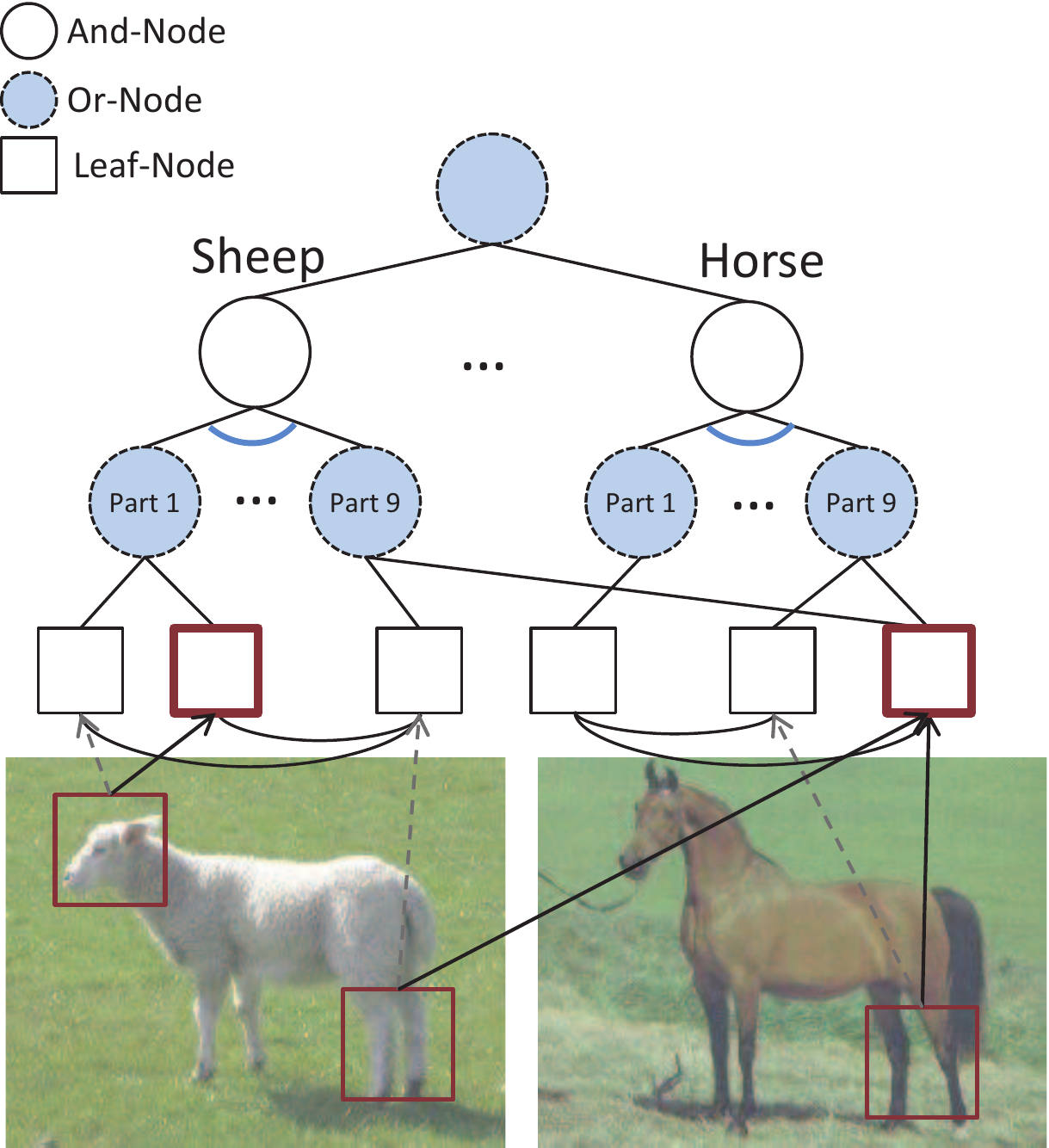,width=2.6in}
\vspace{-1mm}
\caption{An example of the proposed 4-layer And-Or graph model for multiclass object recognition. Parts of the model for sheep and horse are shown. The squares in bottom represent the leaf-nodes, which can be shared among different classes(e.g. the leaf-node for localizing legs are shared between sheep and horse). The or-nodes over bottom are used to activate their children leaf-nodes, tackling the appearance variability. }\label{fig:ModelFig}
\end{figure}

\vspace{-2mm}

Object recognition is an area of active research in computer vision, and its performance has been improved substantially in recent years~\cite{LatentSVM,ConSVM,MultiComponent,LinPR,ICCV09Marr,HoughForest}. The objective of this work is to develop a novel hierarchical and reconfigurable model for multiclass object recognition, in the form of an And-Or graph representation, as Fig.~\ref{fig:ModelFig} illustrates. We study two following issues that are often ignored or over-simplified in previous works.

{\bf Model reconfigurability.}  One key challenge in object modeling is to capture the large object variation in appearance and view/pose. Some recently proposed deformable part-based models~\cite{LatentSVM,ConSVM} handle this challenge by using hierarchical and contextual compositions, and achieve remarkable progresses. However, the structural configurations of these models are mainly fixed, e.g., the number of part detectors and the ways of composition. Inspired by And-Or graph models in~\cite{LinAoG,LeoAOG,NIPS09AOG,HierarchicalPoslets}, we develop the ``switch variables'', namely or-nodes, to specify alternative compositions in hierarchy. In detection, the or-nodes are used to activate its children leaf-nodes (i.e. local classifiers), accounting for intraclass variance. It worths mentioning that the association of or-nodes with its children leaf-nodes can be automatically determined in model training. In Fig.~\ref{fig:ModelFig}, the sheep head is localized by the leaf-node that is activated by its parent or-node.

{\bf Model sharing.} In the context of multiclass object recognition, existing systems commonly treat different classes as unrelated entities. According to acknowledged studies~\cite{JointBoosting,hLDA,HoughForest}, sharing information among different classes can boost model performance in general and alleviate the requirement of a large amount of training data. Recently, Salakhutdinov et al.~\cite{ShareMit} propose a learning-to-share framework that allows rare objects to borrow statistical strength from other related classes, and demonstrate impressive results. It inspires us to make structure shared in the And-Or graph model, for adapting the task of multiclass recognition. In our method, the leaf-nodes are sharable among different classes so that we keep the model compact to represent multiple object categories. For example, in Fig.~\ref{fig:ModelFig}, the part of feet in category horse and sheep have similar appearances, and thus can be  both detected by the leaf-node shared across the two classes.

The key contribution of this work is a novel And-Or graph model for multiclass object recognition, by addressing the both above issues. Without loss of generality, we define our four layered model, as Fig.\ref{fig:ModelFig} illustrates. The leaf-nodes (denoted by squares) in the bottom are discriminative classifiers for detecting object parts.  The or-nodes (denoted by dashed circles) over in the third layer are used to activate one of its children leaf-nodes in detection, which are allowed to slightly perturb for capturing deformations. The and-nodes (denoted by solid circles) in the second layer are global classifiers for object classes. The root-node at top is for switching multiclass recognition, which is also an or-node. In addition, we define the collaborative edges (denoted by curve connections) to encode intraclass (part-level) relations, and interclass contexts are modeled in the similar way as the edges connect the and-nodes also.

One non-trivial problem in model training is to automatically determine the model structure without requiring elaborate supervision and initialization. In our method, we propose a novel algorithm for this problem, namely Dynamical Structural Optimization (DSO), motivated by the recently proposed structural optimization methods~\cite{CCCP,DCCCP}. It is designed in the EM-type iterating with three steps. (i) Estimate model latent variables for optimization, according to parameters from the previous iteration. (ii) Reconfigure the model structure by clustering. In this step, we produce leaf-nodes associated with their parent or-nodes and make leaf-nodes shared across classes. (iii) Check the acceptance for the newly generated model structure, and update the model parameters.

Due to large variance among classes, it would be intractable to train the classes altogether by pooling all samples from different classes into a bag. In this work, we first partition all classes into several groups by a data-driven approach, in order to reduce the computational complexity for model sharing. Then we train the models for object classes in each group. For example, we can easily decide to put sheep and horses into one group and train the multiclass model by sharing. Afterwards, the trained models for all groups are further combined into the complete one, by reweighing parameters of all the models. And the collaborative edges are also learned during this step.

\vspace{-1mm}


\section{Related Work}

\vspace{-1mm}

Traditional multiclass object detectors are trained in a one-vs-all manner, where each object category are trained independently. These methods often rely on large amount of training data. A pioneer work~\cite{JointBoosting} is proposed to learn shared features among classes and improve the classifier in both effectiveness and efficiency. Opelt et al.~\cite{IncJointBoosting} further incorporate the incremental learning with classifier sharing. To discover hierarchical structures of object categories, the Hierarchical Latent Dirichlet Allocation (hLDA) model is presented in~\cite{hLDA}. The efficiency can be significantly improved by integrating taxonomies with object hierarchy~\cite{Taxonomy2}.

To tackle realistic challenges in object recognition, many deformable part-based methods are developed by latent structural learning recently~\cite{LatentSVM,LeoCCCP,ConSVM}. These models are also extended to multiclass recognition and detection~\cite{ICCV09Marr,ShareMit,HoughForest,SharedDPM}. For example, Razavi et al.~\cite{HoughForest} present the multiclass Hough Forest combing with the part-based models; Desai et al.~\cite{ICCV09Marr} further incorporate the context information into hierarchy, and predict a structured labeling for each image during detections. However, the commonly used part-based models are often defined in a tree structure, whose configurations are fixed during the learning and detection, and may have problems on handling objects with large appearance and structure variations.


And-Or graph models are first proposed for modeling complex visual patterns by Zhu and Mumford~\cite{AOG_SCZHU}. Its general idea, using And/Or nodes to account for structural compositions and variabilities in hierarchy, has been applied in several vision tasks, e.g., human parsing~\cite{HierarchicalPoslets,LeoAOG} and object modeling~\cite{LinAoG}. These approaches often require supervised learning or manually initialization. Fidler et al.~\cite{NIPS09AOG} propose to train the And-Or graph for multiclass shape-based detection in a generative way, and extensively discuss the learning strategies. Motivated by these works, we propose an alternating way to discriminatively train the And-Or graph model for multiclass object recognition, and achieve superior performances.

\vspace{-0.5ex}

\section{And-Or Graph Model}

\vspace{-0.5ex}

Our multiclass object model is constructed in the form of an And-Or graph $\mathcal{G}= (\mathcal{V},\mathcal{E})$, where $\mathcal{V}$ contains three types of nodes, $\mathcal{E}$ represents the collaborative edges. The root-node is indexed as $0$, indicating the switch among classes. The and-nodes are indexed by $r=1,...,m$, each representing one category. For each and-node, there are $9$ or-nodes arranged in a layout of $3 \times 3$ blocks to represent object parts, and we index all the or-nodes as $j = m+1,...,10m$. The leaf-nodes in the fourth layer are indexed by $i=10m+1,...,10m+n$, where $n$ is the leaf-node number dynamically adjusted during training. For notation simplicity, we define that $m^{\prime} = 10m + 1, n^{\prime} = 10m + n$, and $i \in ch(j)$ indexes a child node of node $j$. The details of our model are presented as follows.

\textbf{Sharable Leaf-node}: The leaf-nodes $L_i,i=m^{\prime},...,n^{\prime}$ are local classifiers for object parts, and they can be shared among different classes. Specifically, if a leaf-node is affiliated to the $j$-th or-node, it is also possible to be shared by the or-nodes in other classes indexed by ${j+9 \times k}$, where $k \in \{1,2...\}$. We denote the location of leaf-node $L_i$ as $P_i$, which is determined by its parent or-node activating $L_i$ during inference. The response of $L_i$ is defined as,
\begin{eqnarray}\label{eq:LeafScore}
&& R_{i}^{l}(X,P_i) = \omega_i^l \cdot \phi^l(X,P_i).
\end{eqnarray}

In our implementation, a HOG~\cite{HOG} pyramid is built across different image scales as in ~\cite{LatentSVM}. $\phi^l(X,P_i)$ is the HOG feature extracted from image X at position $P_i$, and $\omega_i^l$ is a parameter vector.

\textbf{Or-node}: The or-nodes $U_j,j=m+1,...,10m$ in the third layer are ``switch'' variables to select (activate) their children. For each leaf-node ${L}_i$, we define an variable $V_i \in \{0,1\}$ to represent the activation during inference. An indicator vector is then composed for each or-node ${U}_j$: $\textbf{V}_{j}=(V_{i_1},V_{i_2},...)$, where $i_k \in ch(j)$ and $ \| \textbf{V}_{j} \| = 1/0 $. Note that $|| \textbf{V}_{j} || = 1$ only when one of the leaf-nodes is activated under ${U}_j$. The response of ${U}_j$ is thus defined as,
\begin{eqnarray}\label{eq:OrScore}
&& R_{j}^{u}(X,P_{j},\textbf{V}_{j}) = \sum_{i \in ch(j)} R_{i}^{l}(X,P_{j}) \cdot V_{i},
\end{eqnarray}
where $P_{j}$ denotes the position of $U_j$, and it is allowed to perturb slightly during inference. We define a feature for object deformation as $\phi^s(P_r,P_{j})=(dx,dy,dx^2,dy^2)$, where $(dx,dy)$ represents the displacement of ${U}_j$ relative to its anchor position that is determined by the position of its parent $P_r$. The response of the deformation is defined as,
\begin{eqnarray}\label{eq:ShapeScore}
&& R_{j}^{s}(P_r,P_{j}) = \omega_j^s \cdot \phi^s(P_r,P_{j}).
\end{eqnarray}
\vspace{-3.5ex}

\begin{figure}[!htb]
\centering
\epsfig{figure=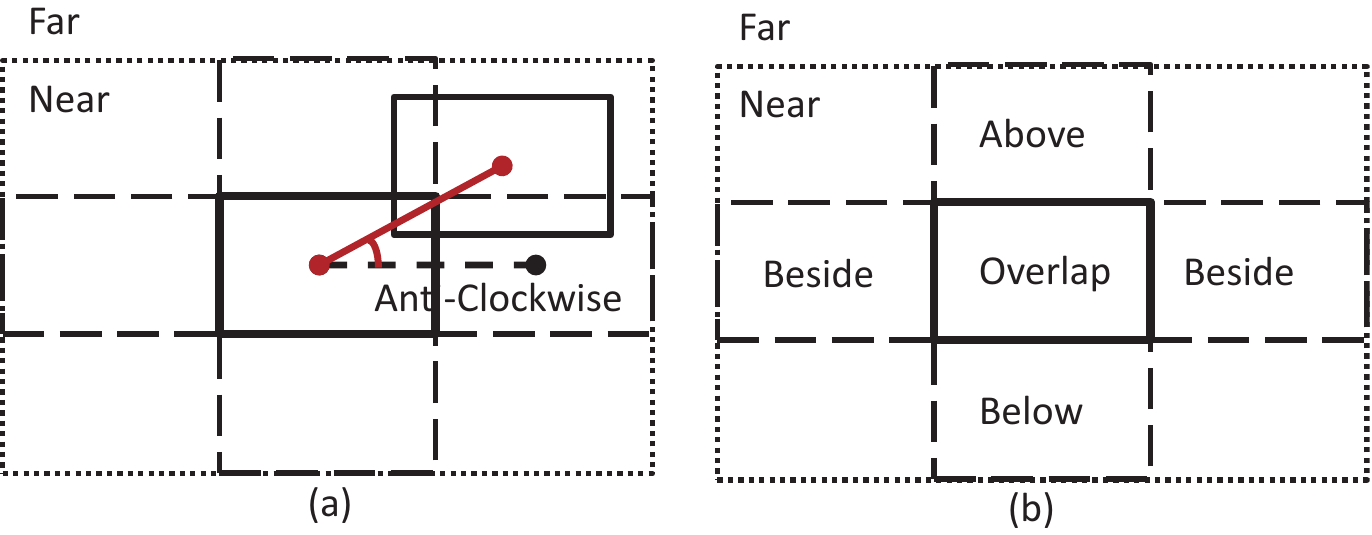,width=0.47\textwidth}
\vspace{-1.2ex}
\caption{Illustration of the features for defining collaborative edges. (a) shows feature $\psi^{l}(P_{i},P_{i^{\prime}})$ between leaf-nodes; (b) shows feature $\psi^{a}(P_r,P_{r^{\prime}})$ between and-nodes.
}
\label{fig:ContextFea}
\end{figure}

\textbf{And-node}: The and-nodes $A_r,r=1,...,m$ are global classifiers for objects. Suppose $A_r$ is placed at $P_r$ during detection, we extract the HOG feature for the and-node $\phi^a(X,P_r)$ at half the resolution of the feature extracted for leaf-nodes. We define the response for $A_r$ with its parameters $\omega_r^a$, as,

\vspace{-3.5ex}
\begin{eqnarray}\label{eq:AndScore}
&& R_{r}^{a}(X,P_r) = \omega_r^a \cdot \phi^a(X,P_r).
\end{eqnarray}
\vspace{-3ex}

\textbf{Root-node}: The root-node on the top is an or-node for switching different classes, i.e. choosing its children and-nodes. Similarly with defining the or-nodes, for each and-node $A_r$, we also define the activation for it as $V_r\in \{0,1\}$, and the indicator vector for root-node is $\textbf{V}_{0}=(V_1,...,V_m)$ and $\| \textbf{V}_{0} \| = 1$, i.e, only one children is selected.

\textbf{Collaborative edge}: There are two types of collaborative edges in our model, representing the spatial co-occurrence between different leaf-nodes as well as between different and-nodes. For the collaborative edges between leaf-nodes, we introduce a 4-bin binary feature $\psi^{l}(P_{i},P_{i^{\prime}})$. Each bin of $\psi^{l}(P_{i},P_{i^{\prime}})$ represents one of the relations: \emph{clockwise, anti-clockwise, near} and \emph{far} between two leaf-nodes $L_i$ and $L_{i^{\prime}}$. As Fig.~\ref{fig:ContextFea}(a) illustrates, the bold rectangle in the middle represents the location of $L_i$. If the center of $L_{i^{\prime}}$ is localized in the dotted rectangle, it is \emph{near} the $L_i$, otherwise it is \emph{far} from $L_i$. We connect the initial centers of $L_i$ and $L_{i^{\prime}}$ with the dashed line and the red line represents their layout after accounting for deformation. Then we use two bins to indicate either \emph{clockwise} or \emph{anti-clockwise} for the angle between the dashed line and the red line. We thus define the response of the collaborative edge between two leaf-nodes as,

\vspace{-3ex}
\begin{eqnarray}\label{eq:EdgeScore2}
&& \Gamma_{i,i^{\prime}}^{l}(P_{i},P_{i^{\prime}}) = \alpha_{i,i^{\prime}}^{l} \cdot \psi^{l}(P_{i},P_{i^{\prime}}),
\end{eqnarray}
where $\alpha_{i,i^{\prime}}^{l}$ is a 4-bin parameter vector. Motivated by~\cite{ICCV09Marr}, we define a 6-bin binary feature $\psi^{a}(P_r,P_{r^{\prime}})$ representing the contextual relations: \emph{above, below, beside, overlap, near} and \emph{far} between two objects. As Fig.~\ref{fig:ContextFea}(b) illustrates, the bold rectangle in the middle represents the window of $A_r$. And the dashed and dotted rectangles represent the bins to be set as $1$ if the center of $A_{r^{\prime}}$ is inside. The response of the collaborative edge between two and-nodes is defined as,
\begin{eqnarray}\label{eq:EdgeScore1}
&& \Gamma_{r,r^{\prime}}^{a}(P_r,P_{r^{\prime}}) = \alpha_{r,r^{\prime}}^{a} \cdot \psi^{a}(P_r,P_{r^{\prime}}),
\end{eqnarray}
where $\alpha_{r,r^{\prime}}^{a}$ is a 6-bin parameter vector. In practice, we only connect the two leaf-nodes whose parent or-nodes are adjacent to each other in spatial domain. And the and-nodes are connected across classes.

\vspace{-1.0ex}

\section{Inference}
\vspace{-0.5ex}

Given an image, the task for inference is to localize all the multiclass objects with the model. For simplicity, we notate the vector of selections for and-nodes together with leaf-nodes as $\mathbb{V}=\langle V_1,...,V_m,V_{m^{\prime}},...,V_{n^{\prime}} \rangle$, and the vector of placements as $\mathbb{P}=\langle P_1,...,P_{10m} \rangle$.

A subgraph of the And-Or graph, rooted at one of the and-nodes, can be regarded as a detector for one class. For each subgraph, we compute its scores by sliding the detection sub-window at different positions and scales of the image. It is a procedure integrating the local testing and binding testing as follows.

\textbf{Local testing}: For a subgraph model rooted at ${A}_r$ (i.e., $V_r = 1$) and placed at $P_r$ of the image, we assume a hypothesis $\mathbb{V}$ for leaf-node selections. Then the placement of each part can be obtained by incorporating Eq.(\ref{eq:OrScore}) and Eq.(\ref{eq:ShapeScore}):

\vspace{-2ex}
\begin{small}
\setlength{\abovedisplayskip}{1pt}
\setlength{\belowdisplayskip}{1pt}
\begin{align}\label{eq:PCal} \widetilde{P}_{j} = & \max_{P_j} (R_{j}^{u}(X,P_{j},\textbf{V}_{j}) - R_{j}^{s}(P_r,P_{j})) \nonumber  \\[-1.5pt]
 = &\max_{P_j}(\sum_{i \in ch(j)} R_{i}^{l}(X,P_{j}) \cdot V_{i} - R_{j}^{s}(P_r,P_{j}) ),
\end{align}
\end{small}
where $R_{i}^{l}(X,P_{j})$ represents the leaf-node response, and we can share these responses among different classes by calculating them at the beginning of inference. Then the score of local testing is calculated  as:

\vspace{-1ex}
\begin{small}
\setlength{\abovedisplayskip}{1pt}
\setlength{\belowdisplayskip}{1pt}
\begin{eqnarray}\label{eq:BotUp}
S_r^{l}(X,\widetilde{\mathbb{P}},\mathbb{V}) = \sum_{j \in ch(r)} (R_{j}^{u}(X,\widetilde{P}_{j},\textbf{V}_{j}) - R_{j}^{s}(P_r,\widetilde{P}_{j})).
\end{eqnarray}
\end{small}

\vspace{-1ex}

\textbf{Binding testing}: We obtain the response over the and-node $R_r^{a}(X,P_r)$ with Eq.(\ref{eq:AndScore}). And for each hypothesis $\mathbb{V}$, we compute the scores of intra-class contextual relations between the selected leaf-nodes via Eq.(\ref{eq:EdgeScore2}). Then the binding score is calculated as:

\vspace{-2.5ex}
\begin{small}
\setlength{\abovedisplayskip}{1pt}
\setlength{\belowdisplayskip}{1pt}
\begin{eqnarray}\label{eq:TopDown}
S_r^{a}(X,\widetilde{\mathbb{P}},\mathbb{V}) = R_r^{a}(X,P_r) + \sum_{\mathclap{i,i^{\prime}=m^{\prime}}}^{n^{\prime}} \Gamma_{i,i^{\prime}}^{l}
(\widetilde{P}_{i},\widetilde{P}_{i^{\prime}}) \cdot V_{i} \cdot V_{i^{\prime}},
\end{eqnarray}
\end{small}
where the leaf-node location is set as $\widetilde{P}_{i} = \widetilde{P}_{j}$ for $i \in ch(j), ||\textbf{V}_{j}|| = 1$. By integrating these two procedures, we select the best $\mathbb{V}$ as the score of detection via the subgraph rooted at ${A}_r$:	

\vspace{-3ex}
\begin{small}
\begin{eqnarray}\label{eq:SubScore}
S^{g}(X,r,P_r) = \max_{\mathbb{V}} (S_r^{l}(X,\widetilde{\mathbb{P}},\mathbb{V}) + S_r^{a}(X,\widetilde{\mathbb{P}},\mathbb{V})).
\end{eqnarray}
\end{small}
\vspace{-3.5ex}

\begin{figure*}
\begin{center}
\epsfig{figure=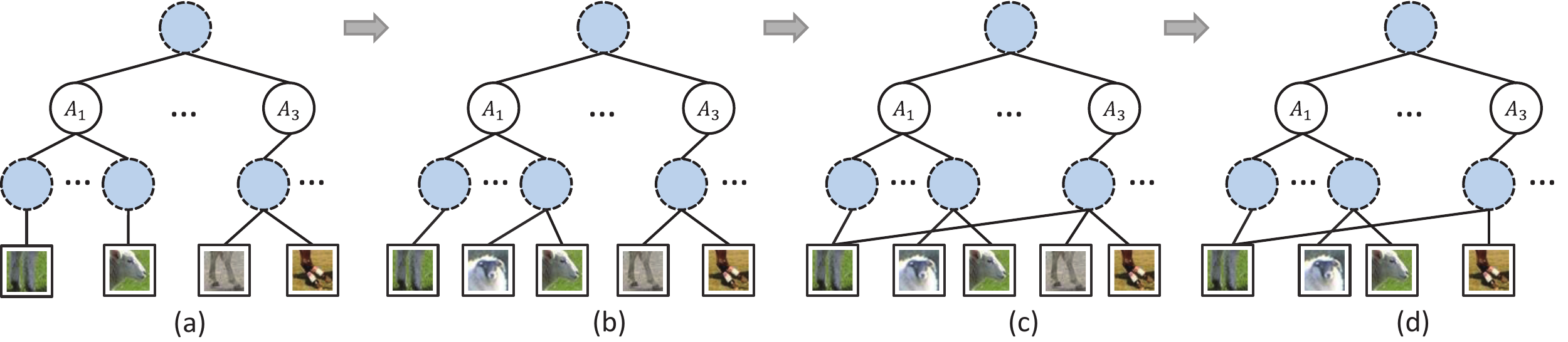,width=0.95\textwidth}
\vspace{-3.5ex}
\end{center}
   \caption{Dynamical Structural Optimization. Parts of the multi-class model for sheep and horse are illustrated in different iterations. (a) The model structure after the first iteration; (b) A new leaf-node is created to recognize the head of sheep; (c) A leaf-node for sheep leg is shared with the horse; (d) A leaf-node for horse leg is removed.}
\label{fig:SDO}
\end{figure*}

After the detections for all subgraphs, we can represent the image as a collection of $K$ scored sub-windows, overlapping at different scales. Our objective is to label them with $Y=\{y_1,...,y_K\}$, where $y_i \in \{1,...,m\}$ represents object classes and $y_i=-1$ the background. The multiclass detection score in the image can be defined by combining  Eq.(\ref{eq:SubScore}) and Eq.(\ref{eq:EdgeScore1}),

\vspace{-1.5ex}
\begin{small}
\setlength{\abovedisplayskip}{1pt}
\setlength{\belowdisplayskip}{1pt}
\begin{eqnarray}\label{eq:FinalScore}
S(X) = \max_{Y} (\sum_{k=1}^K S^{g}(X,y_k,P^k) + \sum_{\mathclap{k,k^{\prime}=1}}^K \Gamma_{y_k,y_{k^{\prime}}}^{a}(P^k,P^{k^{\prime}})),
\end{eqnarray}
\end{small}
where $P^k$ indicates the position of the $k$th sub-window of the image. The optimization of Eq.(\ref{eq:FinalScore}) can be solved by the greedy forward search mentioned in ~\cite{ICCV09Marr}. We define a instance set $\mbox{INS}=\{(k,y)\}$ indicating that $y_k = y$ for $\forall (k,y) \in \mbox{INS}$ and $y_k=0$ otherwise. Then the greedy method is performed as Algorithm \ref{alg:Framwork}.

\begin{small}
\begin{algorithm}[htb]
\caption{Greedy Forward Inference}
\label{alg:Framwork}
\begin{algorithmic}\footnotesize
\REQUIRE ~~\\
    The detections scores $S^{g}(X,r,P_r)$ for image sub-windows.
\ENSURE ~~\\                           
    Instance set $\mbox{INS}$, detection score $S$.

\INPUT ~~\\
  $\mbox{INS}=\{\}$, $S=0$, $\delta(k,y)=S^{g}(X,y,P^k)$ for $k \in \{1,...,K\}$.

\MYWHILE
    \STATE
    \begin{itemize}
\setlength{\itemsep}{1pt}
 \setlength{\parskip}{0pt}
 \setlength{\parsep}{10pt}
      \item[1.] $(\widetilde{k},\widetilde{y})= \mbox{argmax}_{(k,y)} \delta(k,y)$ for $\forall (k,y) \notin \mbox{INS}$.
      \item[2.] $\mbox{INS}=\mbox{INS} \cup (\widetilde{k},\widetilde{y})$.
      \item[3.] $S = S + \delta(\widetilde{k},\widetilde{y})$.
      \item[4.] $\delta(k,y) = \delta(k,y) + \Gamma_{y,\widetilde{y}}^{a}(P^k,{P}^{\widetilde{k}}) + \Gamma_{\widetilde{y},y}^{a}(P^{\widetilde{k}},P^k)$.
    \end{itemize}
\MYENDWHILE {$\delta(\widetilde{k},\widetilde{y}) \leq 0$ and $S$ stops increasing.}

\end{algorithmic}
\end{algorithm}
\end{small}


\vspace{-1.0ex}
\section{Dynamical And-Or Graph Learning}
\vspace{-0.5ex}
The training of our And-Or graph is a two stages procedure: (i) estimating model structure (without edges) and parameters for each object group; (ii) combining models and learning collaborative edges.

To reduce the computational cost for model sharing, we first divide the object classes into several groups as a data-driven initialization, and train the multiclass model for each group. Afterwards, we combine the trained models together to construct the final And-Or graph model.

The learning for stage (i) is an EM-type procedure incorporating structure reconfiguration and parameter estimation. During each iteration, our algorithm dynamically create and remove leaf-nodes associated with their parent or-nodes, and share leaf-nodes among classes. More precisely, a leaf-node is created to better handle the intra-class variance (Fig.~\ref{fig:SDO}(b)); A leaf-node is removed if there is another similar one (Fig.~\ref{fig:SDO}(d)); A leaf-node is shared as it can capture the similar appearances for other classes (Fig.~\ref{fig:SDO}(c)).

\vspace{-0.5ex}
\subsection{Data Driven Initialization}
\vspace{-0.5ex}
Suppose the number of all classes is $M$, we partition them into several groups as a data driven initialization for training. The partition is based on the similarity between two classes, and we calculate the similarity as follows.

(I) We first learn a two-layer deformable part-based model ~\cite{LatentSVM} $\mathcal{T}^k=\{T^k_i\}$ for all classes, where $T^k_i$ represents one part classifier for $k$-th class. And we apply $\mathcal{T}^k$ to perform detection on the positive training samples in every class. During the detection, each $T^k_i$ extracts a set of image patches from different samples, and we group these patches into a cluster $\Omega^k_i$. Note that the size of image patches detected by $T^k_i$ is $(h^k_i,w^k_i)$. For all $\Omega^k_i$, we further merge them into a few new sets, each of which contains image patches of similar size $(h^k_i,w^k_i)$. In each of the new sets, we describe the image patches with the HOG descriptor and group them into several clusters by using ISODATA algorithm with Euclidean distance.

(II) Afterwards, a matrix $\mathcal{M}$ is defined to represent the similarity between $M$ classes. In each set of image patches, if there are patches from class $j$ and $k$ falling into the same cluster, we set $\mathcal{M}(j,k) \leftarrow \mathcal{M}(j,k) + 1$. Two classes $j$ and $k$ are assumed to share their models if $\mathcal{M}(j,k) > \sigma$, where $\sigma$ is a threshold set as $M/3$ empirically.

(III) Based on the calculated $\mathcal{M}$, we assign the classes that are possibly shared into the same group $\mathbb{S}$. We thus obtain a few groups as $\{\mathbb{S}_1,...,\mathbb{S}_c\}$. We denote that each $\mathbb{S}$ has $|\mathbb{S}|$ classes, and we discuss the training method for each $\mathbb{S}$ in the following section.

\vspace{-0.5ex}
\subsection{Optimization Formulation}
\vspace{-0.5ex}
Given an object group $\mathbb{S}$, we train a multiclass model without collaborative edges, which is a procedure integrating structure reconfiguration and parameter estimation. Suppose there are a set of $N$ training samples $(X_1,y_1)$,...,$(X_N,y_N)$ in $\mathbb{S}$, where $X$ is the image, $y \in \{1,...,|\mathbb{S}|\}$ labels the object classes, and $y=-1$ labels the background. At the beginning of training, we initialize the multiclass model with $m=|\mathbb{S}|$ and-nodes and one leaf-node for each or-node. The detection score of this model can be represented as the maximization of Eq.(\ref{eq:SubScore}) over $m$ and-nodes, by setting edge parameters to zero,
\begin{small}
\setlength{\abovedisplayskip}{2pt}
\setlength{\belowdisplayskip}{2pt}
\begin{align}\label{eq:TreeScore}
& S^{t}(X) = \max_{1 \leq r \leq m } S^{g}(X,r,P_r) \nonumber \\[-1.5pt]
& = \max_{\mathbb{P},\mathbb{V}} (\sum_{\mathclap{i=m^{\prime}}}^{n^{\prime}} \omega_i^l \cdot \phi^l (X,P_{i}) \cdot V_{i}
- \sum_{{r=1}}^{m} \sum_{{j \in ch(r)}} \omega_j^s \cdot \phi^s(P_r,P_{j}) \cdot  ||\textbf{V}_{j}||  \nonumber \\[-3pt]
& +  \sum_{r=1}^{m} \omega_r^a \cdot \phi^a (X,P_r) \cdot V_r ),
\end{align}
\end{small}
where the first two terms represent the response of local testings, and the last term is the and-node response. For simplicity, we refer $H=(\mathbb{P},\mathbb{V})$ as the latent variables, then we redefine Eq.(\ref{eq:TreeScore}) in a discriminative form as,
\begin{small}
\setlength{\abovedisplayskip}{2pt}
\setlength{\belowdisplayskip}{2pt}
\begin{align} \label{eq:discriminative_fun}
&& S^{\omega}(X) = argmax_{(y,H)} (\omega \cdot \phi(X,y,H)),
\end{align}
\end{small}
where $\omega$ includes the complete model parameters of current model, $\phi(X,y,H)$ is defined as,
\begin{small}
\setlength{\abovedisplayskip}{1pt}
\setlength{\belowdisplayskip}{1pt}
\begin{equation}\label{eq:feature_phi}
\phi(X,y,H) =
  \left\{
   \begin{array}{lr}
   \phi(X,H) & \mbox{if }   V_y = 1\\
   0 & \mbox{otherwise } \\
   \end{array}
  \right.,
\end{equation}
\end{small}
and $\phi(X,H )$ is the overall feature vector.

The function (\ref{eq:discriminative_fun}) can be learned by applying structural SVM with latent variables,
\begin{small}
\setlength{\abovedisplayskip}{2pt}
\setlength{\belowdisplayskip}{2pt}
\begin{align} \label{eq:learn_opt}
 \min_{\omega} & \frac{1}{2} \| \omega \|^2  +  C\sum_{k=1}^N[\max_{y,H}(\omega \cdot \phi(X_k,y,H) + \mathcal{L}(y_k,y)) \nonumber \\[-2pt]
&- \max_H (\omega \cdot \phi(X_k,y_k,H))],
\end{align}
\end{small}
where $C$ is a penalty parameter set as $0.005$ empirically, and we define the loss function $\mathcal{L}(y_k,y) = 0$ when $y_k = y$, and $\mathcal{L}(y_k,y) = 1$ if $y_k \neq y$. In recent works~\cite{LeoCCCP,LSVMCCCP}, the CCCP~\cite{CCCP} method is applied to solve the non-convex optimization, which provides an iterative approach to achieve a local minima. However, in these methods, the model structure configuration is assumed to be fixed, e.g., without or-nodes. Motivated by these works, we propose a Dynamical Structural Optimization (DSO) method to train out model.

\vspace{-0.5ex}
\subsection{Dynamical Structural Optimization}
\vspace{-0.5ex}

To optimize the objective Eq.(\ref{eq:learn_opt}), we transform it into a concave and convex form following ~\cite{LeoCCCP},

\vspace{-2.5ex}
\begin{small}
\setlength{\abovedisplayskip}{1pt}
\setlength{\belowdisplayskip}{1pt}
\begin{align}\label{eq:cccp_fg}
 \min_{\omega} & [ \frac{1}{2} \|\omega \|^2 + C \sum_{k=1}^N \max_{y,H}(\omega \cdot \phi(X_k,y,H) + \mathcal{L}(y_k,y))]  \nonumber \\[-2pt]
& -  C \sum_{k=1}^N \max_{H} (\omega \cdot \phi(X_k,y_k,H)) \\[-2pt] \label{eq:opt_target}
 =  \min_{\omega} & [f(\omega) - g(\omega)],
\end{align}
\end{small}
where the first two terms in (\ref{eq:cccp_fg})  are represented by $f(\omega)$ and $g(\omega)$ is the other term. Then we present our 3-step Dynamical Structural Optimization method as follows.

{\bf(I)} Suppose we are in the iteration $t$, and $\omega_t$ is the parameter vector updated in the previous iteration. We first find a hyperplane $q_t$ to upper bound $-g(\omega)$ in (\ref{eq:opt_target}),
\begin{equation}\label{eq:CCCPQT}
\setlength{\abovedisplayskip}{2pt}
\setlength{\belowdisplayskip}{2pt}
 -g(\omega) \leq -g(\omega_t) + (\omega-\omega_t) \cdot q_t, \forall \omega.
\end{equation}
\vspace{-2.5ex}

We calculate $q_t$ by finding the optimal latent variables $\widetilde{H}_k = argmax_{H} (\omega_{t} \cdot \phi(X_k,y_k,H))$. That is, we apply the current model to perform detections on the training samples, and the hyperplane is constructed as $q_t = - C\sum_{k=1}^N \phi(X_k,y_k,\widetilde{H}_k)$.

{\bf (II) } We adjust the model by structural reconfiguration and sharing, and it is performed on each one of the $9$ object parts over $m$ classes, independently.  Given a variable vector $\widetilde{H}_k$ for a sample, we can obtain the activation of leaf-nodes and the image patches detected via them. For each leaf-node ${L}_i$, we group the patches detected via it from all samples into a cluster $\Omega_i$, and the size of these patches is $(h_i,w_i)$.

We index the nine object parts by $j (m+1 \leq j \leq m+9)$. For the $j$-th part, we pool the clusters whose corresponding leaf-nodes are associated to or-nodes $U_{j+9k}(0 \leq k < m)$  from $m$ classes together. Then these clusters are further merged into a few new sets, each of which contains patches of similar size $(h_i,w_i)$. For each new set, we describe the image patches with HOG descriptor and perform clustering on them by applying ISODATA with Euclidean distance.

After the clustering, the leaf-nodes are reconfigured as: If a cluster is newly generated, we create a new leaf-node accordingly; we remove a leaf-node if there are few image patches in the corresponding cluster. For a cluster $\Omega_i$, if there are images patches localized by  ${U}_j$(in step{\bf(I)}), we associate the leaf-node ${L}_i$ to ${U}_j$. Thus $L_i$ is shared by different classes for different associations.

\begin{figure}[!htb]
\centering
\epsfig{figure=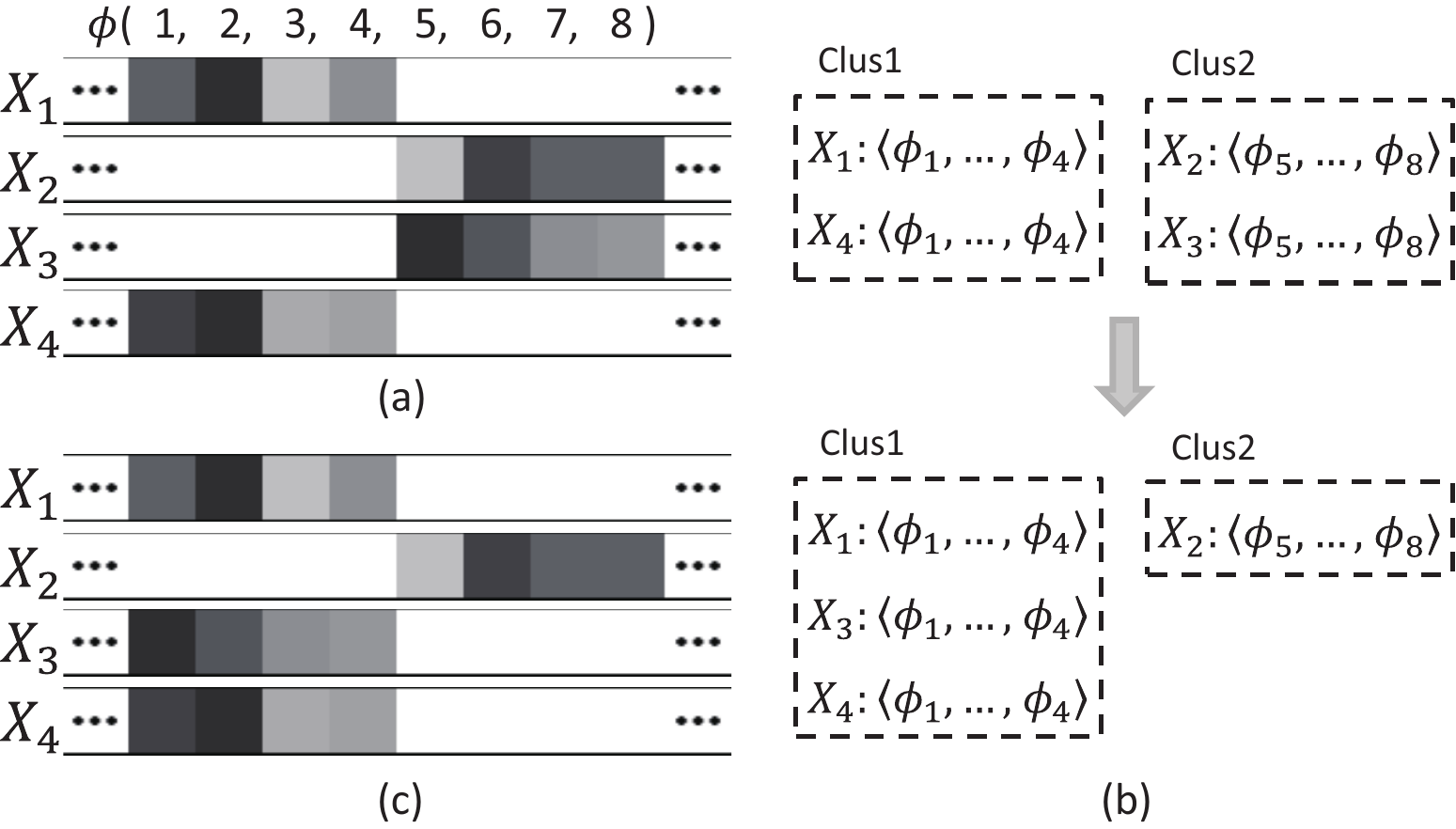,width=0.47\textwidth}
\vspace{-1ex}
\caption{A toy example of feature adjustment according to structural clustering. Parts of 4 feature vectors associated to two different leaf-nodes are presented. (a)shows the feature vectors generated after Step {\bf(I)}, whose value is indicated by the intensities of bins; (b)shows the structural re-clustering: The feature $\langle \phi_5,...,\phi_8 \rangle$ of $X_3$ are moved from Cluster 2 to Cluster 1; (c)updates the feature vectors according to clustering results.}
\label{fig:LearningStep2}
\end{figure}

The feature vector $\phi$ of each sample is also adjusted according to the clustering result. Recall that the HOG vector of an image patch is part of $\phi$, and the patches in the same cluster are represented with the same bins in $\phi$. We present a toy example in Fig.~\ref{fig:LearningStep2} for illustration. The sub-vector $\langle \phi_5,...,\phi_8 \rangle$ of sample $X_3$ is grouped from one cluster to another; then the feature bins are moved from $\langle \phi_5,...,\phi_8 \rangle$ to $\langle \phi_1,...,\phi_4 \rangle$, as (a) and (c) shows. We define the new feature vector for each sample after clustering as $\phi^{d}(X_k,y_k,\widetilde{H}_k)$, then the new hyperplane in step {\bf (I)} is reconstructed as $q_t^d = - C\sum_{k=1}^N \phi^{d}(X_k,y_k,\widetilde{H}_k)$.

{\bf (III) }With the current model structure and $q_t^d$ we can learn the model parameters by solving,

\vspace{-2ex}
\begin{small}
\begin{equation}\label{eq:CCCPS1}
\omega_{t}^{d} = argmin_\omega(f(\omega)+\omega \cdot q_t^d).
\end{equation}
\end{small}
\vspace{-3ex}

By substituting $f(\omega)$ with the first two terms defined in Eq.(\ref{eq:cccp_fg}), we can re-write Eq.(\ref{eq:CCCPS1}) as,

\vspace{-2ex}
\begin{small}
\setlength{\abovedisplayskip}{2pt}
\setlength{\belowdisplayskip}{2pt}
\begin{align} \label{eq:SSVM}
\min_\omega & \frac{1}{2} \|\omega\|^2 +  C\sum_{k=1}^N[\max_{y,H}(\omega \cdot \phi(X_k,y,H) + \mathcal{L}(y_k,y)) \nonumber \\[-2pt]
& -  \omega \cdot \phi^d(X_k,y_k,\widetilde{H}_k)].
\end{align}
\end{small}
\vspace{-2ex}

The optimization of Eq.(\ref{eq:SSVM}) can be solved by standard structural SVM. After that, we can calculate the energy of the objective by $E(\omega_{t}^d) = f(\omega_{t}^d) - g(\omega_{t}^{d})$.

If $E(\omega_{t}^{d}) < E(\omega_{t})$, we accept the new model structure and have $\omega_{t+1} = \omega_{t}^{d}$. Otherwise, we keep the model configuration as it is in the previous iteration, and continue to perform parameter optimization without structure reconfiguration as Eq.(\ref{eq:CCCPS1}): $\omega_{t+1} = argmin_\omega(f(\omega)+\omega \cdot q_t)$.

In this way, we ensure the optimization objective in Eq.(\ref{eq:opt_target}) continuing to decrease in iterations. Thus, the algorithm keeps iterating until the objective converges.

\vspace{-0.5ex}
\subsection{Model Combination}
\vspace{-0.5ex}

After training the multiclass models for each object group in $\{\mathbb{S}_1,...,\mathbb{S}_c\}$, we combine them together into a complete one for all the object categories. Intuitively, the root-nodes from each group are first merged into the final top root-node, so that the original and-nodes are all associated to the new root-node. Then we introduce a $n^{\prime}$ dimension vector $\beta =(\beta_{1},..., \beta_{n^{\prime}})$ to re-weight the parameters for the newly generated model. Meanwhile, the collaborative edges defined in Eq.(\ref{eq:EdgeScore2}) and Eq.(\ref{eq:EdgeScore1}) are trained as well.

For simplicity, we shorten the responses for leaf-node, or-node deformation and and-node as $R_i^l(k)=R_i^l(X,P_{i}^k)$,  $R_j^s(k)=R_j^s(P^{k},P_{j}^k)$ and $R_{y_k}^a=R_{y_k}^a(X,P^k)$. Given an image $X$, the objective function $S(X)$ of multiclass recognition defined in Eq.(\ref{eq:FinalScore}) is reformulated as,

\vspace{-2.5ex}
\begin{small}
\setlength{\abovedisplayskip}{2pt}
\setlength{\belowdisplayskip}{2pt}
\begin{align}\label{eq:Context}
& \max_{Y} \sum_{k=1}^K [ \sum_{{i=m^{\prime}}}^{n^{\prime}} \beta_{i} \cdot R_i^l(k)\cdot V_{i}^k -
\sum_{\mathclap{j \in ch(y_k)}} \beta_{j} \cdot R_j^s(k) + \beta_{y_k} \cdot R_{y_k}^a   \nonumber \\
& +\sum_{\mathclap{i,i^{\prime}=m^{\prime}}}^{n^{\prime}} \alpha_{i,i^{\prime}}^{l} \cdot \psi^{l}(P_{i}^k,P_{i^{\prime}}^k) \cdot V_{i}^k V_{i^{\prime}}^k + \sum_{\mathclap{k^{\prime}=1}}^{K} \alpha_{y_k,y_{k^{\prime}}}^{a} \cdot \psi^{a}(P^{k},P^{k^{\prime}})], \nonumber
\end{align}
\end{small}
where the first two terms represent the local testing score, the next two represent the binding testing score, and the last one accounts for edge responses between and-nodes.

For the training, we collect a set of images containing multiclass objects, each of which is labeled with $Y=\{y_1,...,y_K\}$.  Given each image, we first obtain the latent variables $\widetilde{H}_k$ with Eq.(\ref{eq:discriminative_fun}) by fixing $y_k$, and the responses for each part are derived meanwhile. We can then use the responses $R_i^l(k),  R_j^s(k)$ and $R_{y_k}^a$ as part of the input feature, and train the parameters $\beta$, $\alpha^{l}$ and $\alpha^{a}$ by standard structural SVM. Here the loss function for training is defined as $\mathcal{L}^{\prime}(Y,Y^{\prime})=\mathcal{K}-tp$, where $\mathcal{K}$ indicates the number of objects in groundtruth $Y$, and $tp$ is the number of true positives in $Y^{\prime}$ according to $Y$.

Afterwards, the parameters for leaf-nodes ($A_i,i=m^{\prime},...,n^{\prime}$), or-node deformations($U_j,j=m+1,...,10m$) and and-nodes($A_r,r=1,...,m$) are re-weighed as: $\omega_i^l = \beta_{i} \cdot \omega_i^l$, $\omega_j^s = \beta_{j} \cdot \omega_j^s$ and $\omega_r^a = \beta_{r} \cdot \omega_r^a$.

\vspace{-1ex}

\section{Experiments}

\vspace{-0.5ex}

We evaluate our method on two challenging datasets: UIUC people~\cite{UIUCHuman} and PASCAL VOC 2007~\cite{PASCAL}.

{\em Dataset and Setting.} The UIUC people dataset contains 593 images(296 for training, 297 for testing), and most of them contain one person playing badminton. For PASCAL VOC 2007 dataset, there are 9963 images of $20$ object categories with 5011 images for training and 4952 images for testing. In both datasets, we represent each object category with two views, i.e. each object category is specified by two and-nodes in our model. Hence, we perform 2-class recognition on UIUC people dataset, and 40-class recognition on PASCAL VOC 2007 dataset. During evaluation, we adopt PASCAL Challenge criterion: a detection is considered as correct only if the intersection over union with the groundtruth bounding-box is at least $50\%$. All our experiments are carried out on a PC with Core Duo 3.0 GHZ CPU and 16GB memory. We denote our fully implemented model as ``Ours(full)'', since we will simplify the model in different settings for empirical study.


\vspace{-0.5ex}
\subsection{Experimental Results}

\vspace{-0.5ex}

\begin{figure*}
\begin{center}
\epsfig{figure=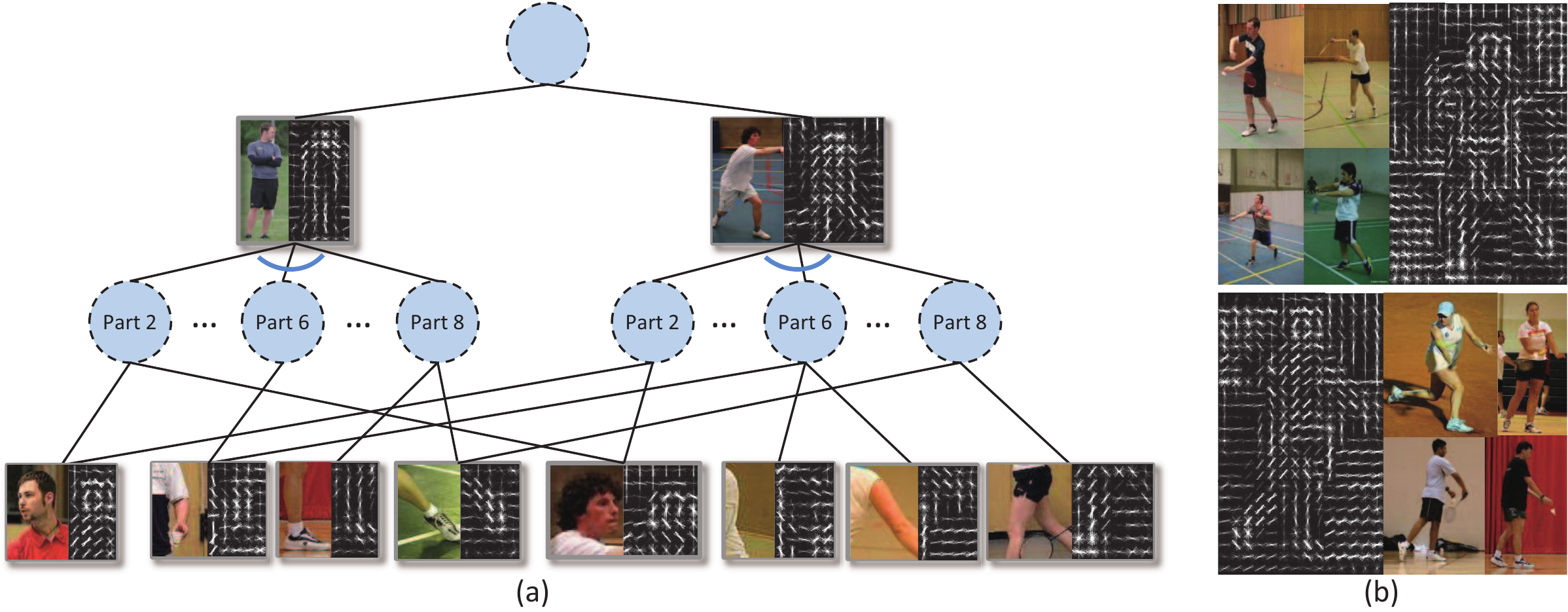,width=0.85\textwidth}
\vspace{-3.4ex}
\end{center}
   \caption{ Visualization of the trained model on UIUC people dataset. (a) shows parts of the model with two classes (views), in which we visualize the detectors (in the form of HOG patterns) for the and-nodes and leaf-nodes based on the learned parameters, alone with the example images recognized by the model. (b) visualizes two detectors that are composed by $9$ activated leaf-nodes. The two detectors are generated, respectively, when recognizing the images beside them. The detectors for and-nodes are not visualized here.}\label{fig:EXP_Human}
\end{figure*}

{\em UIUC people dataset.} For model training, it takes $11$ iterations and around $6$ hours to converge in optimization. And the time for detection on a image is about $5$ seconds. We compare our model with the state-of-the-arts human detectors ~\cite{HierarchicalPoslets,HumanExp1,HumanExp2,LatentSVM}, some of which used manually labeled model. The detection accuracy is calculated as~\cite{HierarchicalPoslets}: only the detection with the highest score on the image is considered. As Table.~\ref{tab:table1} reports, our approach reaches the detection accuracy of $84.5\%$, outperforming other methods. Moreover, we demonstrate the advantage of our model for handing object variations in detection in Fig.~\ref{fig:EXP_Human}. We visualize the detectors generated by our trained model in the form of HOG patterns.  The detectors for and-nodes and leaf-nodes are shown in Fig.~\ref{fig:EXP_Human} (a). Note that some of the leaf-nodes are shared for capturing similar appearances. Two detectors, composed by $9$ activated leaf-nodes, are visualized in Fig.~\ref{fig:EXP_Human} (b). The two detectors are generated when recognizing the images beside them. The results show that our model can generate alterable detectors to adapt diverse object appearances and poses.

\begin{table}	 \small
	\begin{tabular*}{0.48\textwidth}{p{1cm}p{1.1cm}p{0.9cm}p{0.65cm}p{0.65cm}p{0.65cm}p{0.65cm}} \toprule
             &  Ours(full) & Ours(sim) & ~\cite{HierarchicalPoslets} & ~\cite{HumanExp1} & ~\cite{LatentSVM} & ~\cite{HumanExp2} \\
    \hline
    Accuracy & \textbf{0.845} & 0.818 & 0.668 & 0.506 & 0.486  & 0.458
    \\ \bottomrule
    \end{tabular*}
	\caption{Detection accuracies on UIUC people dataset. }\label{tab:table1}
\end{table}

\begin{figure}[!htb]
\centering
\epsfig{figure=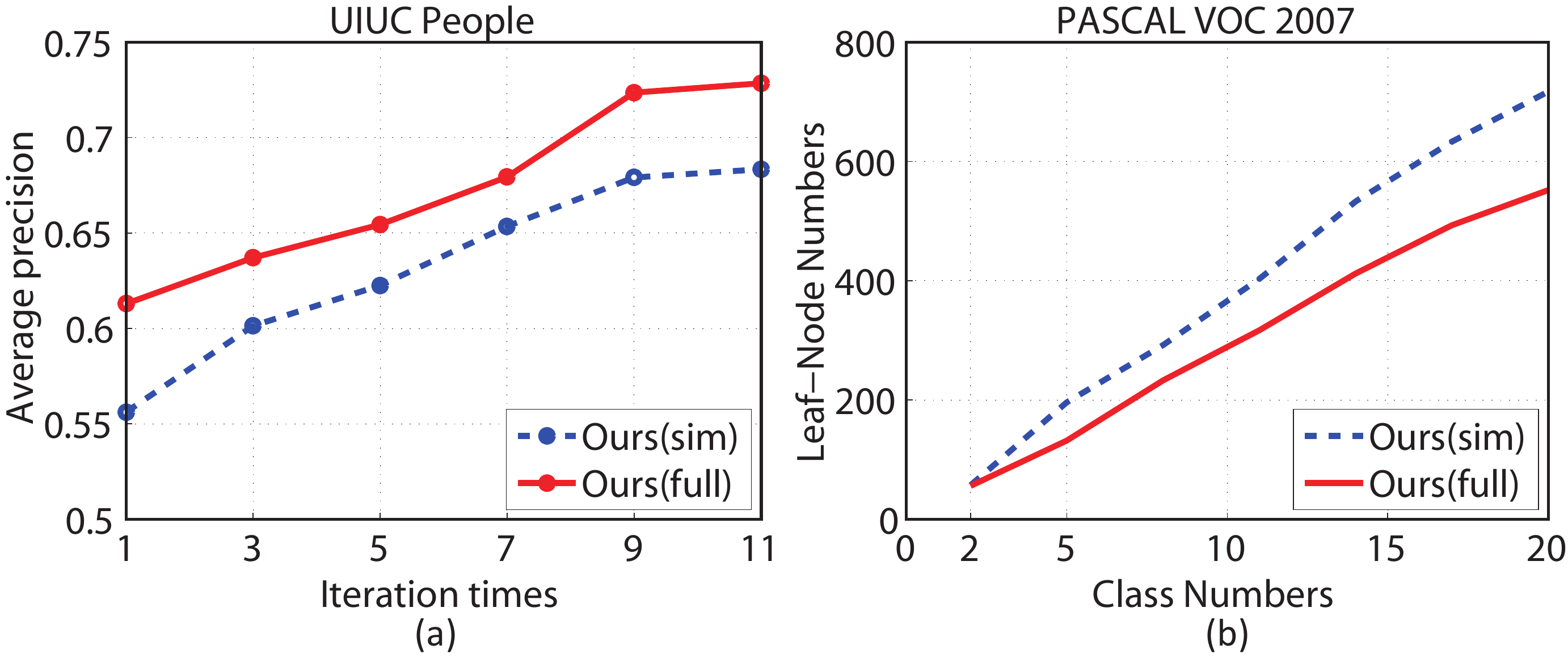,width= 0.48 \textwidth}
\vspace{-3ex}
\caption{Extensive experiments for discussion. ``Our-S'' indicates a simplified model without sharing leaf-nodes. (a) shows the APs on UIUC people dataset. (b) represents the leaf-node numbers with the increasing of object categories on PASCAL VOC 2007 dataset.
}\label{fig:UIUCAP}
\end{figure}

{\em PASCAL VOC 2007 dataset.} To train the 40-class model on the database, it takes $25\sim 30$ iterations in $30\sim 34$ hours. On average, it takes $92$ seconds for detecting all 20 classes of objects on one input image. We then calculate the average precision (AP) to evaluate our method. As shown in Table.~\ref{tab:VOCAP}, our method achieves the mean AP(mAP) of $34.7\%$, which is highly competitive to the state-of-the-arts methods: $29.0\%$~\cite{MultiComponent}, $29.2\%$~\cite{HoughForest}, $29.6\%$~\cite{LeoCCCP}, $32.1\%$~\cite{MKL} and $26.8\%$~\cite{LatentSVM}. We also notice that there is a significant improvement achieving mAP of $37.7\%$~\cite{ConSVM} recently, by employing multi-kernels classification into detection.


\begin{table*} \scriptsize
  \begin{tabular*}{\textwidth}{p{0.9cm}p{0.35cm}p{0.35cm}p{0.35cm}p{0.35cm}p{0.35cm}p{0.35cm}p{0.35cm}p{0.35cm}p{0.35cm}p{0.35cm}p{0.35cm}
  p{0.35cm}p{0.35cm}p{0.35cm}p{0.35cm}p{0.35cm}p{0.35cm}p{0.35cm}p{0.35cm}p{0.35cm}p{0.35cm} lccccccccccccccccccccc}
    \toprule
     & plane
     & bicycle
     & bird
     & boat
     & bottle
     & bus
     & car
     & cat
     & chair
     & cow
     & table
     & dog
     & horse
     & mbike
     & person
     & plant
     & sheep
     & sofa
     & train
     & tv
     & Avg.\\
    \hline
    Ours(full) & 32.5 & \textbf{60.1} & 11.1 & \textbf{16.0} & \textbf{31.0} & \textbf{50.9} & \textbf{59.0} & 26.1 & \textbf{21.2} & 26.5 & 25.4 & 16.4 & \textbf{61.7} & \textbf{48.3} & \textbf{42.2} & \textbf{16.1} & {28.2} & 30.1 & 44.6 & 46.3 & \textbf{34.7}\\

%

    MC~\cite{MultiComponent}    & 33.4 & 37.0 &  15.0 &15.0 & 22.6 & 43.1 & 49.3 & \textbf{32.8} & 11.5 & \textbf{35.8} & 17.8 & 16.3 & 43.6 & 38.2 & 29.8 & 11.6 & \textbf{33.3} & 23.5 & 30.2 & 39.6 & 29.0\\

    HF~\cite{HoughForest}    & 26.0 & 56.0 &  10.0 &11.0 & 21.0 & 47.0 & 50.0 & 16.0 & 19.0 & 23.0 & 20.0 & 12.0 & 51.0 & 45.0 & 37.0 & 12.0 & 17.0 & 29.0 & 41.0 & 38.0 & 29.2\\

    LEO~\cite{LeoCCCP}   & 29.4 & 55.8 &  9.4 & 14.3 & 28.6 & 44.0 & 51.3 & 21.3 & 20.0 & 19.3 & 25.2 & 12.5 & 50.4 & 38.4 & 36.6 & 15.1 & 19.7 & 25.1 & 36.8 & 39.3 & 29.6\\

    MKL~\cite{MKL}   & \textbf{37.6} & 47.8 & \textbf{15.3} & 15.3 & 21.9 & 50.7 & 50.6 & {30.0} & 17.3 & {33.0} & 22.5 & \textbf{21.5} & 51.2 & 45.5 & 23.3 & 12.4 & 23.9 & 28.5 & \textbf{45.3} & \textbf{48.5} & 32.1\\

    UoCTTI~\cite{LatentSVM} &29.0 & 54.6 &  0.6 & 13.4 & 26.2 & 39.4 & 46.4 & 16.1 & 16.3 & 16.5 & 24.5 &  5.0 & 43.6 & 37.8 & 35.0 &  8.8 & 17.3 & 21.6 & 34.0 & 39.0 & 26.8\\

    \bottomrule
  \end{tabular*}
  \caption{Results on PASCAL VOC 2007.}\label{tab:VOCAP}
\end{table*}

\vspace{-0.5ex}

\subsection{Evaluation for Model Sharing}

\vspace{-0.5ex}
To analyze the effectiveness of sharing leaf-nodes, we disable the process for model sharing in training so that we obtain the simplified non-sharing And-Or graph model, named ``Ours(sim)''. As Table.~\ref{tab:table1} reports, ``Ours(sim)'' achieve detection accuracy of $81.8\%$, $2.7\%$ less than the fully implemented model. We also compare the APs of these two models in Fig.~\ref{fig:UIUCAP} (a), in which the APs are visualized with the increasing of iteration numbers for model training. Each AP for a specific iteration number is obtained by testing the model that is trained by the number of iterations.  And the APs of ``Ours(full)'' and ``Ours(sim)'' achieve $72.8\%$ and $68.3\%$, respectively, after $11$ iterations.


We consider the model complexity, represented by the number of leaf-nodes, could be effectively reduced by the model sharing. Thus, we also present an experiment to show the numbers of leaf-nodes in model training, with the increasing of object categories, in Fig.~\ref{fig:UIUCAP}(b). Precisely, we obtain $552$ leaf-nodes for $20$ object categories on PASCAL VOC 2007 dataset, less than $717$ leaf-nodes by ``Ours(sim)'' model.

%

\vspace{-1.5ex}
\section{Conclusion}
\vspace{-1.5ex}

This paper introduces a novel method for multiclass object detection and recognition, in the form of And-Or graph. Our model is shown to handle well the challenges in large variance object recognition. Moreover, we also illustrate the benefits of information sharing among classes, which leads to a more compact and better model. Since our learning method(SDO) is very general, it can be extended to many other vision tasks.

\vspace{-1ex}
\bibliographystyle{ieee}

\end{document}